\documentclass[10pt, a4paper]{article}

\usepackage{lrec-coling2024} % this is the new style
\usepackage{booktabs}
\usepackage{bbding}

\title{Revisiting the MIMIC-IV Benchmark: Experiments Using Language Models for Electronic Health Records}

\name{Jesus Lovon-Melgarejo, Thouria Ben-Haddi, Jules Di Scala,\\
{\bf \large Jose G. Moreno and Lynda Tamine}}  

\address{University of Toulouse, IRIT, 31000 Toulouse, France\\
         \{firstname.lastname\}@irit.fr\\}

\abstract{
The lack of standardized evaluation benchmarks in the medical domain for text inputs can be a barrier to widely adopting and leveraging the potential of natural language models for health-related downstream tasks.  
This paper revisited an openly available MIMIC-IV benchmark for electronic health records (EHRs) to address this issue. First, we integrate the  MIMIC-IV data within the Hugging Face \textit{datasets} library to allow an easy share and use of this collection. Second, we investigate the application of templates to convert EHR tabular data to text. Experiments using fine-tuned and zero-shot LLMs on the mortality of patients task show that fine-tuned text-based models are competitive against robust tabular classifiers. In contrast, zero-shot LLMs struggle to leverage EHR representations. This study underlines the potential of text-based approaches in the medical field and highlights areas for further improvement.%thus, opening opportunities to enlarge the use and evaluation of PLMs for health-related downstream tasks using our proposed revisited MIMIC-IV benchmark. 
\\ \newline }

\begin{document}

\maketitleabstract

\section{Introduction}
% PLMs provide encode textual representations that are transferable to different NLP tasks with optimal results. 
% In the recent year, PLMs have increased in terms of parameters size and training data. 
% The emergence of ChatGPT setup an important milestone, showing that generative-text models represent an important tool for researchers and general public.
% Particularly, in the medical domain, there has been extensive studies to understand the limit and usability of these models for patient and practiciens side, such as interpreting diagnosis and laboratories exams, or determine faster profiles of patients.

% Results showed that while these models perform better QA based tasks than diagnostic tasks of interpeting EHR. An explanation of this limitation is the mismatch of  tabular data and text. Therefore, a better representation of EHR data could improve the performance for patient and practicien usage.
% ...

%%%%%%%%%
Recent advancements in natural language processing (NLP) and information retrieval (IR) tasks have been significantly driven by Transformers-based models, such as BERT  \cite{devlin2018bert} and RoBERTa \cite{liu2019roberta}. These models have been trained on raw linguistic information with minimal supervision. 
Furthermore, the emergence of large language models (LLMs), 
 such as ChatGPT \cite{achiam2023gpt} and 
Llama 2 \cite{touvron2023llama}, has extended these capabilities by scaling in parameters size and training data.
% such as %Flan-T5 \cite{chung2022scaling} and 
% Llama 2 \cite{touvron2023llama}, has extended these capabilities, demonstrating outstanding performance across a diverse set of tasks and domains.
In the medical domain, applying LLMs has emerged as a novel tool for patients and healthcare practitioners \cite{mesko2023imperative}. For example, electronic health records (EHR), composed of non-linguistic information such as laboratory measurements, procedures, and medication codes, are translated into linguistic reports using these models \cite{van2023clinical}. 
However, it is still unclear how useful EHR model representations are in non-linguistic tasks. 
Beyond privacy concerns, the critical issue preventing the broad adoption of LLMs in for this problem is effectively transforming patient structured information from the raw EHR format to a linguistic unstructured format that can leverage the potential of LLMs' text-based representations. 
Existing Transformer-based models for patient data, such as TransformEHR \cite{yang2023transformehr} and BEHRT \cite{li2020behrt}, have adapted their architecture to consider tabular input data. However, this process requires a costly pre-training step that does not take advantage of the advancements in improved LLMs and free EHR benchmarks such as MIMIC IV \cite{johnson2023mimic}. The latter provides large-scale intensive care unit (ICU) patient data in a tabular form related to established cohorts used in different downstream tasks (e.g., mortality patient classification).
Consequently, we argue that improving the accessibility of these resources to meet the models' evolution is crucial for the field.

In this paper, we propose a simple but effective methodology to standardize the MIMIC-IV benchmark towards using state of the art (SOTA) Transformer-based architectures (BERT, DistilBERT \cite{Sanh2019DistilBERTAD} and RoBERTa), and LLMs (Llama 2, Meditron \cite{chen2023meditron70b}) for health-related predictive tasks. 
For this purpose, we identify six main groups of features on the ICU data and propose a template-based data-to-text transformation. Thus, we are able to provide a text document input that summarizes the patient's ICU entry. Additionally, and for the sake of reproducibility, we provide a Hugging Face \textit{datasets} object\footnote{\scriptsize\url{https://huggingface.co/docs/datasets/index}} that automatically produces a clinical cohort in the desired textual format\footnote{Publicly available at \url{https://huggingface.co/datasets/thbndi/Mimic4Dataset}}. Our main contributions are as follows: 1) A standard MIMIC-IV benchmark, integrated into the Hugging Face \textit{datasets} library%\footnote{Publicly available at https://huggingface.co/datasets/xxxxx/Mimic4Dataset}
, allowing flexible use of the EHRs representation in health-related downstream tasks;   2) A comprehensive set of experiments using eight different models for evaluating the effectiveness of our revisited MIMIC-IV benchmark on the mortality classification task.

\section{Background and Related Work}

\subsection{MIMIC Collections and Benchmarks}
The Medical Information Mart for Intensive Care (MIMIC) collection \cite{johnson2023mimic,mimic} is one of the largest and most recent EHR datasets. It includes more than 250,000 patients admitted to intensive care at Beth Israel Deaconess at Boston's Beth Israel Deaconess Medical Center. For each patient, details of the full journey in the ICU are available  in a deidentified form for privacy concerns\footnote{With regard to the Safe Harbor provision of the HIPAA.% (Health Insurance Portability and Accountability Act).
}. 
The current version is the MIMIC-IV collection \cite{gupta2022extensive}
%, which is an updated version of the MIMIC-III \cite{johnson2016mimic} collection 
which collect patient data between 2008-2019 and uses ICD-9 and ICD-10 versions of the International Classification of Diseases (ICD)\footnote{\scriptsize\url{https://www.who.int/standards/classifications/classification-of-diseases}} to list diagnoses and to link medical procedures to diagnoses.\\ 
%\vspace{-0.5cm}
%\subsubsection{MIMIC benchmarks.}
In recent works, multiple benchmarks were proposed for the medical domain \cite{Harutyunyan2019,gupta2022extensive,wang2020mimic} using MIMIC collections \cite{johnson2023mimic,johnson2016mimic}. They appear as a mainstream mean of model comparability and reproducibility.
%solution to the impossibility to compare different models when complete details of previous work where missing despite the fact that all where using the same common resource. 
%One of the first well known benchmarks was proposed by Harutyunyan et al. \cite{Harutyunyan2019}  based on the MIMIC-III collections.\\
%collection and ensure the use of a predefined cohort group, same procedure for feature construction, coherent number of features, and fixed partitions of training and test, in order to facilitate the replication and comparison to previous results. 
The MIMIC-IV data pipeline \cite{gupta2022extensive} is proposed to preprocess data for downstream tasks. This pipeline is able to transform raw data into a ready-to-use tabular representation of the patient's data. Additionally, it provides the mapping to ICD as well as standard techniques for dimensionality reduction. 
Although a first step is the proposal of the benchmarks, we aim to go for two steps forward in this work by proposing the integration of the MIMIC IV  benchmark into \textit{datasets} of Hugging Face\footnote{Our implementation respects MIMIC's access policies by asking the user to provide the original data.}, one of the largest hub ready-to-use datasets, as well as the possibility %of use text for the mortality classification task.
of using Transformer-based models (including LLMs) for predictive tasks on EHRs.

\subsection{Transformers for EHRs}
Transformers-based models of the general domain, such as BERT, have been adapted to the clinical domain using medical-related linguistic collections such as PubMed (BioBERT \cite{lee2020biobert} and ClinicalBERT \cite{alsentzer2019publicly}). Recently, efforts to encode non-linguistic information of EHRs to model patient data have emerged with models such as BEHRT \cite{li2020behrt}, Med-BERT \cite{rasmy2021med}, and TransformEHR \cite{yang2023transformehr}.
These models encode different health modalities in flexible architectures. However, they require pre-training on large-scale datasets and do not benefit from the significant progress of Transformer-based models in the NLP literature.
Furthermore, LLMs such as ChatGPT \cite{achiam2023gpt}, Llama 2 \cite{touvron2023llama}, and its medical variant Meditron \cite{chen2023meditron70b} have shown outstanding performance in different clinical tasks related to adapting non-linguistic health data, such as images and EHR diagnostics, into text \cite{mesko2023imperative, yeo2023assessing}. However, the exploration of this linguistic EHR representation for non-linguistic tasks, referred to as EHR downstream tasks, is limited. In order to bridge this gap, we present experiments on EHR data to explore their potential.

%\subsection{Transformer models for medical data}
\section{MIMIC-IV Benchmark Revisited}
Here, we detail the pipeline and EHR data used, then we describe the templates proposed for transforming tabular EHR data into textual inputs.
\subsection{The Pipeline}
\begin{figure*}[h]
%\vspace{-0.1cm}
\caption{Dataset generation pipeline for the tabular format and the text format.\label{ourpipeline}}
\centering
\includegraphics[width=1\textwidth]{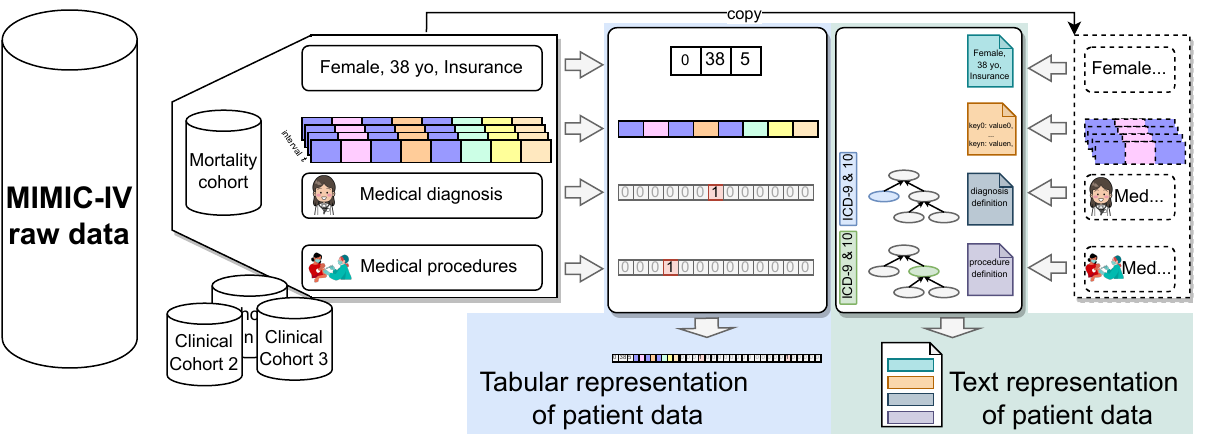}
\end{figure*}

We rely on the MIMIC-IV benchmark to produce the standard evaluation framework for text. Thus, first, we integrated the recommended pre-processing guidelines in the \textit{datasets} library and implemented all the features of the MIMIC-IV-Data-Pipeline\footnote{\scriptsize\url{https://github.com/healthylaife/MIMIC-IV-Data-Pipeline}} provided in a tabular form, as shown in the left side of Figure \ref{ourpipeline}.  After the preprocessing steps, we obtained a tabular representation that includes the demographic, current diagnosis features  and time-series features related to labs, medications, procedures, and vitals, as show in Table \ref{features}.

{
\begin{table}
\caption{Features list for the MIMIC-IV benchmark.}
\label{features}
\begin{tabular}{p{1.3cm}p{5.4cm}}
\toprule
Name & Description\\ \midrule
\small \textit{Demo graphics (DEMO)}:& \small The list of demographic data is a tiny vector corresponding to the patient's gender, ethnicity, medical insurance, and age category. This data is encoded to obtain a numerical vector.\\ 
\small \textit{Diagnosis (COND)}:& \small The list of diagnoses established on a patient's admission is encoded using a one-hot vector of all ICD codes %\cite{ICD10} \\
including the patient's identified diseases. Note that this vector could be large w.r.t. other features.\\% large size (near to 1,000) depending of the desired cohort.
 \small \textit{Chart Events/Lab (CHART/ LAB)}:& \small Gives the value of the biological \textit{item\_id} performed in time interval \textit{t}.\\
  \small\textit{Medications (MEDS)}:&\small For each \textit{item\_id} corresponding to a medication the quantity administered in time interval \textit{t} or zero if not administrated.\\
  \small\textit{Procedures (PROC)}:&\small The list of medical procedures performed is given as a form of a one-hot vector setting to 1 the \textit{item\_id} of procedures performed in time interval \textit{t}.\\
  \small\textit{Output Events (OUTE)}:&\small The list of biological samples taken is encoded using a one-hot vector of each \textit{item\_id} of the samples performed in time interval \textit{t}.\\
\bottomrule
\end{tabular}
\end{table}
}

 {\small
\begin{table*}[!t]
\caption{Example of a text-based representation of a patient from the MIMIC-IV benchmark dataset. Values were changed to avoid leaking the example. \label{exemple}}
\begin{center}
\resizebox{\textwidth}{!}{%
\begin{tabular}{cl}
\toprule

Feature  & Example text                                                                                         \\ \midrule
DEMO     & The patient white male, 55 years old, covered by Other                                           \\
COND     & was diagnosed with Streptococcal sepsis; Acute pancreatitis; resistance to anti-microbial drugs.                                 \\
CHAR/LAB & The chart events measured were: 73.655 for Heart Rate; 116.859 for Heart rate Alarm - High; ...  \\
MEDS     & The mean amounts of medications administered during the episode were: 44.778 of Albumin 5\%; ... \\
PROC     & The procedures performed were: Dialysis Catheter; 18 Gauge; EKG; ...                             \\
OUTE      & The outputs collected were: OR EBL; OR Urine; Pre-Admission; ...     \\
\bottomrule
\end{tabular}
}
\end{center}
\end{table*}
}
%\vspace{-1.0cm}
Note that features like CHAR/LAB are given in time intervals, thus a reduction/expansion strategy must be applied to normalize the representation size. Data imputation is commonly applied by sampling data from a fixed number of time windows or even averaging values across a sequence of time windows. As shown in Section \ref{tabresults}, we did not find large differences between sampling or averaging these features.

\subsection{Proposed Templates}

Finally, feature EHR data is transformed to text using a template-based strategy as shown in right side of Figure \ref{ourpipeline} and described below: 

``\textit{The patient \{ethnicity\} \{gender\}, \{age\} years old, covered by \{insurance\}
was diagnosed with \{cond\_text\}.}'' With 
%demographic data and
\{cond\_text\} corresponding to the textual description from ICD10 of the diagnoses.

``\textit{The chart events measured were: \{chart\_text\}.}'' With {chart\_text} the list of
biological measurements of the form: \{mean\_val\} for \{feat\_label\}, mean\_val is the
mean value of the \{feat\_label\} measurement over the episode.

``\textit{The mean amounts of medications administered during the episode were:
\{meds\_text\}.}'' With \{meds\_text\} the list of quantities of drugs administrated
of the form: \{mean\_val\} for \{feat\_label\}, \{mean\_val\} the average value
over the episode of the quantity of drug \{feat\_label\}.

``\textit{The procedures performed were: \{proc\_text\}.}'' With \{proc\_text\} the list of
medical procedures performed during the episode.

``\textit{The outputs collected were: \{out\_text\}.}'' With \{out\_text\} the list of biological prebiological samples taken during the episode.

Table \ref{exemple} shows a sample of the textual input.

\section{Experiments}
\subsection{Experimental Setup}
To ensure a fair reproducibility of our experiments, we develop a \textit{datasets} object that is able to produce tabular information as well as template-based textual data. 

\begin{table*}[t]
\caption{Comparative evaluation of  our standardized MIMIC-IV vs. original benchmark \cite{gupta2022extensive} on the patient mortality classification task. \label{tabular-results}}
\begin{center}
\resizebox{\textwidth}{!}{%
\begin{tabular}{ccccc|cc}
\toprule
          & \multicolumn{2}{c}{\textbf{Representation 1}} & \multicolumn{2}{c|}{\textbf{Representation 2}}                             & \multicolumn{2}{c}{\textbf{\cite{gupta2022extensive}}}                               \\ 
\textbf{Algorithm}                                                               & \textbf{AU-ROC}   & \textbf{AU-PRC}  & \textbf{AU-ROC}  & \textbf{AU-PRC}  & \textbf{AU-ROC}                      & \textbf{AU-PRC}                     \\ \midrule
\textbf{Gradient Boosting}                                                     & 0.86              & 0.53             & 0.86             & 0.53             & 0.85                                 & 0.48                                \\
\textbf{XGBoost}                                                               & 0.86              & 0.51             & 0.85             & 0.51             & 0.84                                 & 0.47                                \\
\textbf{Random Forest}                                                         & 0.82              & 0.49             & 0.84             & 0.50             & 0.79                                 & 0.39                                \\
\textbf{Logistic Regression}                                                   & 0.77              & 0.36             & 0.77             & 0.37             & 0.67                                 & 0.24                               \\ \bottomrule
\end{tabular}
%\vspace{-6cm}
}
\end{center}
\end{table*}

For \textbf{tabular data}, we create \textit{Representation 1}, which follows the default configuration used in \cite{gupta2022extensive}, but other configurations are available in our implementation. Similarly, \textit{Representation 2} is an aggregated representation of the same data. The main difference is the number of final features as the former uses 2766 features (as result of the concatenation of each window representation) and the latter 1110 features (as the values of all windows are averaged). We evaluated our revised MIMIC-IV benchmark on patient mortality classification as a pilot downstream task as provided in \cite{gupta2022extensive}. Evaluation focuses on both benchmark reproducibility (Cf. Section 4.2)  and both feasibility and effectiveness using representative models (Cf. Section 4.3).
Model parameters were selected using a 5-fold cross validation for classical machine learning algorithms available on Scikitlearn library\footnote{\scriptsize\url{https://scikit-learn.org/}}. We used algorithms for tabular data, such as Gradient Boosting (default parameters), XGBoost (objective=``binary :logistic''), Random Forest (n\_estimators=$300$, criterion=``gini''), and Logistic regression (default parameters). 

For \textbf{textual data}, we fine-tuned six different Transformer-based models. We used optimal hyperparameters including learning rate of $5e-5$, AdamW optimization and $3$ epochs. For our zero-shot setup with LLMs, we explored multiple prompts. In the following, we report two of these prompts, which provide the highest number of valid responses for the task. We limited the output generation to $2$ tokens.

We refer as \textit{P1} for the prompt:

\vspace{0.6cm}
\noindent\fbox{%
    \parbox{\columnwidth}{%
Prompt P1: ``You are an extremely helpful healthcare assistant. You answer the question using only 'yes' or 'no' and considering a patient hospital profile: `[textual EHR]'. \\
Question: Is the patient dead?. \\
Answer (only yes or no): ''
    }%
}
\vspace{0.3cm}

Similarly, we refer as \textit{P2} for the prompt:

\vspace{0.6cm}
\noindent\fbox{%
    \parbox{\columnwidth}{%
Prompt P2: ``Analyze the provided ICU data for a patient. The data covers the first 48 hours of the ICU stay, including vital statistics, lab test results, and treatments administered. Answer only Yes for a prediction of survival or No for a prediction of mortality. The patient ICU data is: `[textual EHR]'. Based on this data, answer. \\
Question: Will the patient survive in the next 24 hours?. \\
Answer (use only yes or no): ''
    }%
}
\vspace{1cm}

We set a limit of $512$ tokens for input length for fine-tuned models and $1024$ tokens for zero-shot models. It should be noted that this truncation only affected the fine-tuned models, and at times, it removed relevant information related to MEDS, PROC, and OUTE features. In Section 4.3, we discuss an ablation study that looks into the impact of these features.

\subsection{Evaluation with Tabular EHR Data\label{tabresults}}
Our results on tabular data and the reference values from the original benchmark \cite{gupta2022extensive} are presented in Table \ref{tabular-results}. Note that our results are presented for two different aggregation strategies, \textit{Representation 1} and \textit{Representation 2}. In both cases, our results are slightly higher than those of the approach proposed in \cite{gupta2022extensive} and used as a starting point. This is mainly due to our careful pre-processing of the data. As an important result, note that the \textit{Representation 2} column performs similarly to \textit{Representation 1} but uses significantly fewer features. Additionally, 1,034 values among 1,110 from the vector representation are sparse as they are dedicated to the diagnosis representation. These results lead us to pursue the text-based representation as only 66 values from biological signals combined with textual data (diagnosis) are enough to achieve state-of-the-art results on tabular data.

% \end{table}

\subsection{Evaluation of Using Template-based Text Inputs}
Our main results on using text-based models for patient mortality classification tasks are presented in Table \ref{text-results}. For the fine-tuned models, we used the three general purpose trained models, namely DistilBERT (distilbert-base-uncased \cite{Sanh2019DistilBERTAD}), BERT (bert-base-uncased \cite{devlin2018bert}), and RoBERTa (roberta-base \cite{liu2019roberta}) (top three), and three others from the medical domain, namely  BioClinicalBERT (Bio\_ClinicalBERT \cite{alsentzer2019publicly}), BioBERT (dmis-lab/biobert-v1.1 \cite{lee2020biobert}), and BiomedNLP (microsoft/BiomedNLP \cite{pubmedbert}( (bottom three). We reported only results with oversampling\footnote{We found similar results without oversampling.}. Our results show that the general purpose and domain-specific models behave similarly regarding AU-ROC, with all models getting close values (between 0.87 and 0.88). However, AU-PRC values differ as models from the medical domain outperform the general-purpose ones. Although a slight improvement was observed for general-purpose models in terms of AU-PRC, this is not enough to achieve the performance of the domain-specific models. Unsurprisingly, there is a clear interest in fine-tuning medical texts. However, general-purpose models, such as RoBERTa, closely follow top performances. 

Furthermore, we explored using two LLMs, Llama2 (13b) (meta-llama/Llama-2-7b-hf \cite{touvron2023llama}) and its medical variant Meditron (7b) (epfl-llm/meditron-7b \cite{chen2023meditron70b}) in a zero-shot setup considering two different prompts named \textit{P1} and \textit{P2}. We generally observed a lower performance from the Zero-shot section (as shown in Table \ref{text-results}) compared to Fine-tuned models. After analyzing the Zero-shot section, we found that prompt \textit{P1} received better scores than \textit{P2}. These results indicate that models are sensitive to the query format for this task. In addition, we noticed that domain-specific models, such as Meditron, performed better than general ones like Llama 2, using both prompts, similar to the fine-tuned setup. These findings suggest that SOTA LLMs struggle to encode and transfer EHR representations to downstream tasks within the explored prompts. A possible development towards using LLMs with tabular data is to define better translation methods to integrate this structured knowledge into the language models. Also, these findings motivate further research and experimentation by applying alternative techniques such as in-context learning \cite{dongsurvey} or prompt-tuning \cite{lesterpower}.

Moreover, in this setup, in addition to right or wrong answers, we also consider unanswered questions. Such questions occur when the LLM fails to provide an output from the expected tokens, which are ``Yes'' or ``No'' in our case. For our experiments, we consider ``No'' the default answer for the results reported in Table \ref{text-results}. 
To provide further details, we display the number of answered and unanswered questions per model in Table \ref{tab:llms}. Upon analysis, we found that the Llama 2 model left $3.30\%$ of the dataset unanswered, while the Meditron model left only $0.04\%$ unanswered using prompt \textit{P1}. In contrast, the prompt \textit{P2} obtained  $69.37\%$ of unanswered questions with Llama 2 and no unanswered questions by Meditron.

By comparing the different prompts used to describe the task, we can observe that Llama 2 (general domain model) struggles to understand the task while making some modifications. In contrast, Meditron (domain-specific models) is more stable when using different reformulations of the task.

\begin{table}[]
\caption{Results of the general purpose and medical domain models on the patient mortality task using text representations of patient data.\label{text-results}}

\centering
\resizebox{\columnwidth}{!}{%

\begin{tabular}{ccc} \toprule
 %& \multicolumn{4}{c}{\textbf{Mortality task}}  \\
\textbf{Models}                                                 & \textbf{AU-ROC}                & \textbf{AU-PRC}           \\  \midrule
\multicolumn{3}{c}{\scriptsize \textbf{Fine-tuned}}\\
DistilBERT                              & 0.87                           & 0.42                          \\
BERT                                    & 0.87                           & 0.43                          \\
RoBERTa                                         & 0.88                           & 0.47                          \\
BioClinicalBERT                                    & 0.87                           & 0.43                          \\
BioBERT                                & 0.88                           & 0.45                          \\
BiomedNLP                                   & 0.88                           & 0.46                         \\ \midrule
\multicolumn{3}{c}{\scriptsize \textbf{Zero-shot with prompt P1}}\\
%\textbf{Flan-T5-xl}     & 0.50    & 0.05                         \\ 
Llama 2 (13b) &                                  0.50                          & 0.38                         \\ 
Meditron (7b) & 0.61 & 0.39 \\
\multicolumn{3}{c}{\scriptsize \textbf{Zero-shot with prompt P2}}\\
%\textbf{Flan-T5-xl}     & 0.50    & 0.05                         \\ 
Llama 2 (13b) &                                  0.50                          & 0.13                         \\ 
Meditron (7b) & 0.51 & 0.23 \\

\bottomrule
\end{tabular}
}
\end{table}

\begin{table}[]
    \caption{Number of answered and unanswered samples by the LLMs for the zero-shot setup.}
    \centering
    \begin{tabular}{ccc}
    \toprule
    \textbf{Model} & \textbf{\# answered} & \textbf{\# unanswered} \\
    \midrule
    \multicolumn{3}{c}{\scriptsize \textbf{With prompt P1}}\\
    Llama 2 (13b)& 5952 \scriptsize($96.70\%$)& 203 \scriptsize($3.30\%$)\\
    Meditron (7b) &6152 \scriptsize($99.96\%$)& 3 \scriptsize($0.04\%$)\\
    \multicolumn{3}{c}{\scriptsize \textbf{With prompt P2}}\\
    Llama 2 (13b)& 1885 \scriptsize($30.63\%$)& 4270 \scriptsize($69.37\%$)\\
    Meditron (7b) &6155 \scriptsize($100.0\%$)& 0 \scriptsize($0.00\%$)\\
         \bottomrule
    \end{tabular}

    \label{tab:llms}
\end{table}

\begin{table*}[]
\caption{Ablation study using different text features. `\Checkmark' indicates that the feature was used in the patient representation. CH/LA stands for CHART/LAB. Results in bold indicate best performance.\label{ablation-results}}
\begin{center}

\resizebox{0.7\textwidth}{!}{%
\begin{tabular}{llccccccc}
\toprule
                                                     & COND     & \multicolumn{1}{c}{\Checkmark}                       & \multicolumn{1}{c}{\Checkmark}                       & \multicolumn{1}{c}{\Checkmark}                       & \multicolumn{1}{c}{\Checkmark}                       & \multicolumn{1}{c}{\Checkmark}         & \multicolumn{1}{c}{\Checkmark}   \\
                                                     & DEMO  & \multicolumn{1}{l}{}                        & \multicolumn{1}{c}{\Checkmark}                       & \multicolumn{1}{c}{\Checkmark}                       & \multicolumn{1}{c}{\Checkmark}                       & \multicolumn{1}{c}{\Checkmark}         & \multicolumn{1}{c}{\Checkmark} \\
                                                     & CH/LA  & \multicolumn{1}{l}{} & \multicolumn{1}{l}{} & \multicolumn{1}{c}{\Checkmark}                       & \multicolumn{1}{c}{\Checkmark}                       & \multicolumn{1}{c}{\Checkmark}         & \multicolumn{1}{c}{\Checkmark}         & \multicolumn{1}{c}{\Checkmark}                                   \\
                                                     & MEDS   & \multicolumn{1}{l}{} & \multicolumn{1}{l}{} & \multicolumn{1}{l}{} &  \multicolumn{1}{c}{\Checkmark}                       & \multicolumn{1}{c}{\Checkmark}         & \multicolumn{1}{c}{\Checkmark}  \\
                                                     & PROC    & \multicolumn{1}{l}{} & \multicolumn{1}{l}{} & \multicolumn{1}{l}{} & \multicolumn{1}{l}{} & \multicolumn{1}{c}{\Checkmark}         & \multicolumn{1}{c}{\Checkmark}                     \\
\multicolumn{1}{c}{\textbf{Features}}                & OUTE                      & &                      &                      &               &               & \multicolumn{1}{c}{\Checkmark}            \\  \midrule
%\multicolumn{1}{c}{\textbf{Model}}                 &               & \textbf{A-R}         & \textbf{A-P}         & \textbf{A-R}         & \textbf{A-P}         & \textbf{A-R}         & \textbf{A-P}         & \textbf{A-R}         & \textbf{A-P}         & \textbf{A-R}  & \textbf{A-P}  & \textbf{A-R}  & \textbf{A-P}  & \textbf{A-R}          \\ \midrule
\multicolumn{9}{c}{AU-ROC}\\
\multicolumn{1}{c}{BERT} &               & 0.87                 & 0.88           & 0.86           & \textbf{0.89}               & \textbf{0.89}  & 0.88  & 0.75                     \\
\multicolumn{1}{c}{BiomedNLP}       &               & 0.87                 & 0.88                      & \textbf{0.88}               & 0.87                             & 0.86              & 0.88          & 0.63                        \\ \midrule
\multicolumn{9}{c}{AU-PRC}\\
\multicolumn{1}{c}{BERT} &               & 0.41                 & 0.44           & 0.40           & \textbf{0.46}               & \textbf{0.46}  & \textbf{0.46}  & 0.24                     \\
\multicolumn{1}{c}{BiomedNLP}       &               & 0.45                 & 0.46                      & \textbf{0.50}               & 0.42                             & 0.39              & 0.47          & 0.13                        \\
\bottomrule   
\end{tabular}
}
\end{center}
\end{table*}

We further our analysis by performing an ablation study with two representative models, BERT and BiomedNLP, to study the accumulative effect of the features. Results are presented in Table \ref{ablation-results}. As a main feature, we can easily identify COND as a clear buster in performance. This feature alone achieves a close value to top performance, indicating that it is a clear signal of the patient profile representation. However, other no-expert-based features, such as \textit{CHAR/LAB}, are also reliable. Note that this is an encouraging result as the features are given in an aggregated form. In fact, compared to the best model, the model can perform correctly in terms of AU-ROC. Also, note that both models achieve the best performance before integrating all the features. In particular, \textit{MEDS, PROC, and OUTE} (only for BiomedNLP) do not improve previous combinations. This indicates that more elaborated templates are worth investigating to integrate these features.

\section{Conclusion}
In this paper, we presented a publicly available Hugging Face \textit{datasets} object that allows a reproducible way to use the MIMIC-IV benchmark for EHR representation and use in health-related tasks based on text-based model. Using MIMIC IV in the proposed object, we aim to facilitate the experimentation with a comprehensive public EHR dataset in its original tabular form and text format through text-based templates. Our experiments showed that fine-tuned text-based models perform similarly to the strongest tabular-based alternatives regarding AU-ROC. On the contrary, LLMs in a zero-shot setup suggested limitations when encoding EHR information. This evaluation provides a starting point for a new family of large language models toward improving the current SOTA on health-related predictive tasks.

% papers.

% \section{Providing References}

% \subsection{Bibliographical References} 

% Bibliographical references should be listed in alphabetical order at the end of the paper. The title of the section, ``Bibliographical References'', should be a Level 1 Heading. The first line of each bibliographical reference should be justified to the left of the column, and the rest of the entry should be indented by 0.35 cm.

% The examples provided in Section~\ref{sec:reference} (some of which are fictitious references) illustrate the basic format required for papers in conference proceedings, books, journal articles, PhD theses, and books chapters.

% \subsection{Language Resource References}

% Language resource references should be listed in alphabetical order at the end of the paper.

\vspace{3em}

\nocite{*}
\section{Bibliographical References}\label{sec:reference}

\bibliographystyle{lrec-coling2024-natbib}
\bibliography{lrec-coling2024-example}

\begin{thebibliography}{31}
\expandafter\ifx\csname natexlab\endcsname\relax\def\natexlab#1{#1}\fi

\bibitem[{Achiam et~al.(2023)Achiam, Adler, Agarwal, Ahmad, Akkaya, Aleman,
  Almeida, Altenschmidt, Altman, Anadkat et~al.}]{achiam2023gpt}
Josh Achiam, Steven Adler, Sandhini Agarwal, Lama Ahmad, Ilge Akkaya,
  Florencia~Leoni Aleman, Diogo Almeida, Janko Altenschmidt, Sam Altman,
  Shyamal Anadkat, et~al. 2023.
\newblock Gpt-4 technical report.
\newblock \emph{arXiv preprint arXiv:2303.08774}.

\bibitem[{Alsentzer et~al.(2019)Alsentzer, Murphy, Boag, Weng, Jindi, Naumann,
  and McDermott}]{alsentzer2019publicly}
Emily Alsentzer, John Murphy, William Boag, Wei-Hung Weng, Di~Jindi, Tristan
  Naumann, and Matthew McDermott. 2019.
\newblock Publicly available clinical bert embeddings.
\newblock In \emph{Proceedings of the 2nd Clinical Natural Language Processing
  Workshop}, pages 72--78.

\bibitem[{Chen et~al.(2023)Chen, Hernández-Cano, Romanou, Bonnet, Matoba,
  Salvi, Pagliardini, Fan, Köpf, Mohtashami, Sallinen, Sakhaeirad, Swamy,
  Krawczuk, Bayazit, Marmet, Montariol, Hartley, Jaggi, and
  Bosselut}]{chen2023meditron70b}
Zeming Chen, Alejandro Hernández-Cano, Angelika Romanou, Antoine Bonnet, Kyle
  Matoba, Francesco Salvi, Matteo Pagliardini, Simin Fan, Andreas Köpf,
  Amirkeivan Mohtashami, Alexandre Sallinen, Alireza Sakhaeirad, Vinitra Swamy,
  Igor Krawczuk, Deniz Bayazit, Axel Marmet, Syrielle Montariol, Mary-Anne
  Hartley, Martin Jaggi, and Antoine Bosselut. 2023.
\newblock \href {http://arxiv.org/abs/2311.16079} {Meditron-70b: Scaling
  medical pretraining for large language models}.
\newblock \emph{arXiv preprint arXiv:2311.16079}.

\bibitem[{Choi et~al.(2019)Choi, Xu, Li, Dusenberry, Flores, Xue, and
  Dai}]{Choi2019LearningTG}
E.~Choi, Zhen Xu, Yujia Li, Michael~W. Dusenberry, Gerardo Flores, Yuan Xue,
  and Andrew~M. Dai. 2019.
\newblock \href {https://api.semanticscholar.org/CorpusID:210839714} {Learning
  the graphical structure of electronic health records with graph convolutional
  transformer}.
\newblock In \emph{AAAI Conference on Artificial Intelligence}.

\bibitem[{Chung et~al.(2022)Chung, Hou, Longpre, Zoph, Tay, Fedus, Li, Wang,
  Dehghani, Brahma et~al.}]{chung2022scaling}
Hyung~Won Chung, Le~Hou, Shayne Longpre, Barret Zoph, Yi~Tay, William Fedus,
  Yunxuan Li, Xuezhi Wang, Mostafa Dehghani, Siddhartha Brahma, et~al. 2022.
\newblock Scaling instruction-finetuned language models.
\newblock \emph{arXiv preprint arXiv:2210.11416}.

\bibitem[{Devlin et~al.(2018)Devlin, Chang, Lee, and
  Toutanova}]{devlin2018bert}
Jacob Devlin, Ming-Wei Chang, Kenton Lee, and Kristina Toutanova. 2018.
\newblock Bert: Pre-training of deep bidirectional transformers for language
  understanding.
\newblock \emph{arXiv preprint arXiv:1810.04805}.

\bibitem[{Dong et~al.(2022)Dong, Li, Dai, Zheng, Wu, Chang, Sun, Xu, and
  Sui}]{dongsurvey}
Qingxiu Dong, Lei Li, Damai Dai, Ce~Zheng, Zhiyong Wu, Baobao Chang, Xu~Sun,
  Jingjing Xu, and Zhifang Sui. 2022.
\newblock A survey on in-context learning.
\newblock \emph{arXiv preprint arXiv:2301.00234}.

\bibitem[{Gu et~al.(2021)Gu, Tinn, Cheng, Lucas, Usuyama, Liu, Naumann, Gao,
  and Poon}]{pubmedbert}
Yu~Gu, Robert Tinn, Hao Cheng, Michael Lucas, Naoto Usuyama, Xiaodong Liu,
  Tristan Naumann, Jianfeng Gao, and Hoifung Poon. 2021.
\newblock Domain-specific language model pretraining for biomedical natural
  language processing.
\newblock \emph{ACM Transactions on Computing for Healthcare (HEALTH)},
  3(1):1--23.

\bibitem[{Gupta et~al.(2022)Gupta, Gallamoza, Cutrona, Dhakal, Poulain, and
  Beheshti}]{gupta2022extensive}
Mehak Gupta, Brennan Gallamoza, Nicolas Cutrona, Pranjal Dhakal, Raphael
  Poulain, and Rahmatollah Beheshti. 2022.
\newblock \href {https://proceedings.mlr.press/v193/gupta22a.html} {{An
  Extensive Data Processing Pipeline for MIMIC-IV}}.
\newblock In \emph{Proceedings of the 2nd Machine Learning for Health
  symposium}, volume 193 of \emph{Proceedings of Machine Learning Research},
  pages 311--325. PMLR.

\bibitem[{Harutyunyan et~al.(2019)Harutyunyan, Khachatrian, Kale, Ver~Steeg,
  and Galstyan}]{Harutyunyan2019}
Hrayr Harutyunyan, Hrant Khachatrian, David~C. Kale, Greg Ver~Steeg, and Aram
  Galstyan. 2019.
\newblock \href {https://doi.org/10.1038/s41597-019-0103-9} {Multitask learning
  and benchmarking with clinical time series data}.
\newblock \emph{Scientific Data}, 6(1):96.

\bibitem[{Johnson et~al.()Johnson, Bulgarelli, Pollard, Horng, Celi, and
  Mark}]{mimic}
Alistair~EW Johnson, Lucas Bulgarelli, Tom~J Pollard, Steven Horng, L~A Celi,
  and R~Mark.
\newblock \href {https://doi.org/10.13026/6mm1-ek67/} {{MIMIC-IV version 2.2}}.

\bibitem[{Johnson et~al.(2023)Johnson, Bulgarelli, Shen, Gayles, Shammout,
  Horng, Pollard, Hao, Moody, Gow et~al.}]{johnson2023mimic}
Alistair~EW Johnson, Lucas Bulgarelli, Lu~Shen, Alvin Gayles, Ayad Shammout,
  Steven Horng, Tom~J Pollard, Sicheng Hao, Benjamin Moody, Brian Gow, et~al.
  2023.
\newblock Mimic-iv, a freely accessible electronic health record dataset.
\newblock \emph{Scientific data}, 10(1):1.

\bibitem[{Johnson et~al.(2016)Johnson, Pollard, Shen, Lehman, Feng, Ghassemi,
  Moody, Szolovits, Anthony~Celi, and Mark}]{johnson2016mimic}
Alistair~EW Johnson, Tom~J Pollard, Lu~Shen, Li-wei~H Lehman, Mengling Feng,
  Mohammad Ghassemi, Benjamin Moody, Peter Szolovits, Leo Anthony~Celi, and
  Roger~G Mark. 2016.
\newblock Mimic-iii, a freely accessible critical care database.
\newblock \emph{Scientific data}, 3(1):1--9.

\bibitem[{Kale and Rastogi(2020)}]{kale2020text}
Mihir Kale and Abhinav Rastogi. 2020.
\newblock Text-to-text pre-training for data-to-text tasks.
\newblock \emph{arXiv preprint arXiv:2005.10433}.

\bibitem[{Lee et~al.(2020)Lee, Yoon, Kim, Kim, Kim, So, and
  Kang}]{lee2020biobert}
Jinhyuk Lee, Wonjin Yoon, Sungdong Kim, Donghyeon Kim, Sunkyu Kim, Chan~Ho So,
  and Jaewoo Kang. 2020.
\newblock Biobert: a pre-trained biomedical language representation model for
  biomedical text mining.
\newblock \emph{Bioinformatics}, 36(4):1234--1240.

\bibitem[{Lester et~al.(2021)Lester, Al-Rfou, and Constant}]{lesterpower}
Brian Lester, Rami Al-Rfou, and Noah Constant. 2021.
\newblock The power of scale for parameter-efficient prompt tuning.
\newblock \emph{arXiv preprint arXiv:2104.08691}.

\bibitem[{Li et~al.(2020)Li, Rao, Solares, Hassaine, Ramakrishnan, Canoy, Zhu,
  Rahimi, and Salimi-Khorshidi}]{li2020behrt}
Yikuan Li, Shishir Rao, Jos{\'e} Roberto~Ayala Solares, Abdelaali Hassaine,
  Rema Ramakrishnan, Dexter Canoy, Yajie Zhu, Kazem Rahimi, and Gholamreza
  Salimi-Khorshidi. 2020.
\newblock Behrt: transformer for electronic health records.
\newblock \emph{Scientific reports}, 10(1):7155.

\bibitem[{Liu et~al.(2019)Liu, Ott, Goyal, Du, Joshi, Chen, Levy, Lewis,
  Zettlemoyer, and Stoyanov}]{liu2019roberta}
Yinhan Liu, Myle Ott, Naman Goyal, Jingfei Du, Mandar Joshi, Danqi Chen, Omer
  Levy, Mike Lewis, Luke Zettlemoyer, and Veselin Stoyanov. 2019.
\newblock Roberta: A robustly optimized bert pretraining approach.
\newblock \emph{arXiv preprint arXiv:1907.11692}.

\bibitem[{MacAvaney et~al.(2021)MacAvaney, Yates, Feldman, Downey, Cohan, and
  Goharian}]{macavaneysigir2021-irds}
Sean MacAvaney, Andrew Yates, Sergey Feldman, Doug Downey, Arman Cohan, and
  Nazli Goharian. 2021.
\newblock Simplified data wrangling with ir\_datasets.
\newblock In \emph{SIGIR}.

\bibitem[{Mesk{\'o} and Topol(2023)}]{mesko2023imperative}
Bertalan Mesk{\'o} and Eric~J Topol. 2023.
\newblock The imperative for regulatory oversight of large language models (or
  generative ai) in healthcare.
\newblock \emph{NPJ digital medicine}, 6(1):120.

\bibitem[{Rasmy et~al.(2021)Rasmy, Xiang, Xie, Tao, and Zhi}]{rasmy2021med}
Laila Rasmy, Yang Xiang, Ziqian Xie, Cui Tao, and Degui Zhi. 2021.
\newblock Med-bert: pretrained contextualized embeddings on large-scale
  structured electronic health records for disease prediction.
\newblock \emph{NPJ digital medicine}, 4(1):86.

\bibitem[{Romanov and Shivade(2018)}]{romanov2018lessons}
Alexey Romanov and Chaitanya Shivade. 2018.
\newblock Lessons from natural language inference in the clinical domain.
\newblock In \emph{Proceedings of the 2018 Conference on Empirical Methods in
  Natural Language Processing}, pages 1586--1596.

\bibitem[{Sanh et~al.(2019)Sanh, Debut, Chaumond, and
  Wolf}]{Sanh2019DistilBERTAD}
Victor Sanh, Lysandre Debut, Julien Chaumond, and Thomas Wolf. 2019.
\newblock Distilbert, a distilled version of bert: smaller, faster, cheaper and
  lighter.
\newblock \emph{ArXiv}, abs/1910.01108.

\bibitem[{Solaiman et~al.(2019)Solaiman, Brundage, Clark, Askell, Herbert-Voss,
  Wu, Radford, Krueger, Kim, Kreps et~al.}]{solaiman2019release}
Irene Solaiman, Miles Brundage, Jack Clark, Amanda Askell, Ariel Herbert-Voss,
  Jeff Wu, Alec Radford, Gretchen Krueger, Jong~Wook Kim, Sarah Kreps, et~al.
  2019.
\newblock Release strategies and the social impacts of language models.
\newblock \emph{arXiv preprint arXiv:1908.09203}.

\bibitem[{Song et~al.(2017)Song, Rajan, Thiagarajan, and Spanias}]{song2018}
Huan Song, Deepta Rajan, Jayaraman~J. Thiagarajan, and Andreas Spanias. 2017.
\newblock Attend and diagnose: Clinical time series analysis using attention
  models.
\newblock In \emph{Proceedings of the 2018 AAAI Association for the Advancement
  of Artificial Intelligence}.

\bibitem[{Touvron et~al.(2023)Touvron, Martin, Stone, Albert, Almahairi,
  Babaei, Bashlykov, Batra, Bhargava, Bhosale et~al.}]{touvron2023llama}
Hugo Touvron, Louis Martin, Kevin Stone, Peter Albert, Amjad Almahairi, Yasmine
  Babaei, Nikolay Bashlykov, Soumya Batra, Prajjwal Bhargava, Shruti Bhosale,
  et~al. 2023.
\newblock Llama 2: Open foundation and fine-tuned chat models.
\newblock \emph{arXiv preprint arXiv:2307.09288}.

\bibitem[{Van~Veen et~al.(2023)Van~Veen, Van~Uden, Blankemeier, Delbrouck,
  Aali, Bluethgen, Pareek, Polacin, Reis, Seehofnerova
  et~al.}]{van2023clinical}
Dave Van~Veen, Cara Van~Uden, Louis Blankemeier, Jean-Benoit Delbrouck, Asad
  Aali, Christian Bluethgen, Anuj Pareek, Malgorzata Polacin, Eduardo~Pontes
  Reis, Anna Seehofnerova, et~al. 2023.
\newblock Clinical text summarization: Adapting large language models can
  outperform human experts.
\newblock \emph{Research Square}.

\bibitem[{Vaswani et~al.(2017)Vaswani, Shazeer, Parmar, Uszkoreit, Jones,
  Gomez, Kaiser, and Polosukhin}]{vaswani2017attention}
Ashish Vaswani, Noam Shazeer, Niki Parmar, Jakob Uszkoreit, Llion Jones,
  Aidan~N Gomez, {\L}ukasz Kaiser, and Illia Polosukhin. 2017.
\newblock Attention is all you need.
\newblock \emph{Advances in neural information processing systems}, 30.

\bibitem[{Wang et~al.(2020)Wang, McDermott, Chauhan, Ghassemi, Hughes, and
  Naumann}]{wang2020mimic}
Shirly Wang, Matthew~BA McDermott, Geeticka Chauhan, Marzyeh Ghassemi,
  Michael~C Hughes, and Tristan Naumann. 2020.
\newblock Mimic-extract: A data extraction, preprocessing, and representation
  pipeline for mimic-iii.
\newblock In \emph{Proceedings of the ACM conference on health, inference, and
  learning}, pages 222--235.

\bibitem[{Yang et~al.(2023)Yang, Mitra, Liu, Berlowitz, and
  Yu}]{yang2023transformehr}
Zhichao Yang, Avijit Mitra, Weisong Liu, Dan Berlowitz, and Hong Yu. 2023.
\newblock Transformehr: transformer-based encoder-decoder generative model to
  enhance prediction of disease outcomes using electronic health records.
\newblock \emph{Nature Communications}, 14(1):7857.

\bibitem[{Yeo et~al.(2023)Yeo, Samaan, Ng, Ting, Trivedi, Vipani, Ayoub, Yang,
  Liran, Spiegel et~al.}]{yeo2023assessing}
Yee~Hui Yeo, Jamil~S Samaan, Wee~Han Ng, Peng-Sheng Ting, Hirsh Trivedi, Aarshi
  Vipani, Walid Ayoub, Ju~Dong Yang, Omer Liran, Brennan Spiegel, et~al. 2023.
\newblock Assessing the performance of chatgpt in answering questions regarding
  cirrhosis and hepatocellular carcinoma.
\newblock \emph{Clinical and molecular hepatology}, 29(3):721.

\end{thebibliography}

%\section{Language Resource References}
%\label{lr:ref}
%\bibliographystylelanguageresource{lrec-coling2024-natbib}
%\bibliographylanguageresource{languageresource}

\end{document}